\documentclass[10pt,a4paper]{article}
\usepackage[utf8]{inputenc}
\usepackage[T1]{fontenc}
\usepackage{amsmath}
\usepackage{amssymb}
\usepackage{graphicx}
\usepackage{hyperref}
\usepackage{xcolor}
\usepackage{booktabs}
\usepackage{multirow}
\usepackage{subfig}
\usepackage{authblk}
\usepackage{enumitem}
\usepackage[left=2.5cm,right=2.5cm,top=2.5cm,bottom=2.5cm]{geometry}

\title{K-means Enhanced Density Gradient Analysis for Urban and Transport Metrics Using Multi-Modal Satellite Imagery}

\author[1,2]{Paweł Tomkiewicz}
\author[2]{Jacek Jaworski}
\author[2]{Paweł Zielonka}
\author[1]{Antoni Wilinski}
\affil[1]{WSB Merito University Gdańsk, Poland}
\affil[2]{NexRI Laboratory for Artificial Intelligence, Gdańsk, Poland}

\begin{document}

\maketitle

\begin{abstract}
This paper presents a novel computational approach for evaluating urban metrics through density gradient analysis using multi-modal satellite imagery, with applications including public transport and other urban systems. By combining optical and Synthetic Aperture Radar (SAR) data, we develop a method to segment urban areas, identify urban centers, and quantify density gradients. Our approach calculates two key metrics: the density gradient coefficient ($\alpha$) and the minimum effective distance (LD) at which density reaches a target threshold. We further employ machine learning techniques, specifically K-means clustering, to objectively identify uniform and high-variability regions within density gradient plots. We demonstrate that these metrics provide an effective screening tool for public transport analyses by revealing the underlying urban structure. Through comparative analysis of two representative cities with contrasting urban morphologies (monocentric vs polycentric), we establish relationships between density gradient characteristics and public transport network topologies. Cities with clear density peaks in their gradient plots indicate distinct urban centers requiring different transport strategies than those with more uniform density distributions. This methodology offers urban planners a cost-effective, globally applicable approach to preliminary public transport assessment using freely available satellite data. The complete implementation, with additional examples and documentation, is available in an open-source repository under the MIT license at \url{https://github.com/nexri/Satellite-Imagery-Urban-Analysis}.
\end{abstract}

\section{Introduction}
Recent advancements in remote sensing technology have greatly expanded access to diverse satellite data types, including optical imagery and Synthetic Aperture Radar (SAR). Platforms such as the Copernicus Data Space, Sentinel Hub, and commercial providers have democratized these datasets for researchers, urban planners, and commercial users \cite{drusch2012sentinel, torres2012gmes, berger2012esa}. This proliferation of open-access satellite data has catalyzed numerous analytical methodologies in urban studies, agriculture, environmental monitoring, and disaster management \cite{pesaresi2016operating, ban2015global}.

However, effectively analyzing this vast repository of multi-modal satellite data presents significant challenges, requiring robust computational resources and advanced mathematical modeling capabilities. Traditional methods that rely on manual processing and simplistic modeling often fall short in fully exploiting satellite data's potential \cite{wu2003estimating, small2016humans}. In response, an ecosystem of startups and research initiatives has emerged to bridge the gap between complex satellite data and actionable insights, with particular focus on SAR data processing, analytics automation, and user-friendly dissemination \cite{geoawesomeness2024geospatial}. Despite these advancements, there remains a critical need for simple, robust, and computationally efficient methodologies suitable for resource-limited environments where traditional, expensive urban planning methods may not be feasible or cost-effective \cite{esch2017breaking}.

A pressing global challenge is optimizing transportation infrastructure, which underpins economic growth, urban sustainability, and social equity. In many urban areas, transportation networks — including communications pathways, logistics, and especially public transit systems — lag behind optimal efficiency, contributing to congestion, pollution, and reduced quality of life \cite{bertaud2018order, rodrigue2020geography}. Addressing this challenge requires innovative approaches capable of rapidly and reliably assessing transportation infrastructure efficiency through spatial urban characteristics.

This paper proposes a novel computational approach designed to analyze urban density gradients using multi-modal satellite imagery for public transport evaluation. Our methodology integrates optical and SAR data to efficiently segment urban areas, identify urban centers, and quantify spatial density distributions. We introduce two key metrics — the density gradient coefficient ($\alpha$) and the minimum effective distance (LD) — offering urban planners a straightforward yet powerful means of preliminary assessment that directly connects urban morphology to public transport optimization strategies. This approach provides significant advantages in accessibility, resource utilization, and scalability, enabling robust transportation infrastructure assessment for diverse urban configurations, including monocentric and polycentric city structures \cite{angel2012dimensions, newman2015end}.

The urban density gradient, which describes how building and population density change with distance from urban centers, has significant implications for public transport planning. Steep gradients (high $\alpha$ values) typically indicate compact urban forms that efficiently support mass transit, while shallow gradients suggest sprawling development patterns requiring different transport solutions. Through our analysis, we demonstrate that cities can be categorized based on their gradient density plots — with clear distinctions between monocentric cities (showing a single dominant peak) and polycentric cities (exhibiting multiple density peaks). These morphological differences have direct implications for optimal public transport network design.

\section{Related Work}

\subsection{Urban Density Gradient Models and Transport Planning}
Urban density gradient analysis originated with the classic monocentric city model pioneered by Alonso \cite{alonso1964location}, Muth \cite{muth1969cities}, and Mills \cite{mills1972studies}. This foundational work established the negative exponential density function describing how population density typically decreases with distance from the city center. Clark's \cite{clark1951urban} empirical work validated this model across numerous cities, confirming that population density follows the pattern:

\begin{equation}
D(r) = D_0 e^{-\alpha r}
\end{equation}

Where:
\begin{itemize}
\item $D(r)$ is the density at distance r from the center
\item $D_0$ is the central density
\item $\alpha$ is the density gradient coefficient
\end{itemize}

Bertaud and Malpezzi \cite{bertaud2003spatial} expanded this analysis to a global dataset of 48 cities, demonstrating how density gradients vary across different urban contexts and development stages, highlighting the relationship between density patterns and transportation efficiency. Cervero and Kockelman \cite{cervero1997travel} further developed the connection between urban form and transport planning through their ``3Ds'' framework—density, diversity, and design—to explain how built environment characteristics influence travel behavior. Ewing and Cervero \cite{ewing2010travel} later expanded this to the ``5Ds'' by adding destination accessibility and distance to transit, firmly establishing the theoretical link between urban density patterns and transport efficiency.

\subsection{Monocentric versus Polycentric Urban Development}
While the monocentric model provides a useful baseline, contemporary urban forms often exhibit polycentric structures. Garreau \cite{garreau1991edge} documented the emergence of ``edge cities,'' while Gordon and Richardson \cite{gordon1996beyond} tracked the systematic decentralization of employment and commercial activities in metropolitan areas. Theoretical frameworks for understanding polycentric development have been advanced by Anas et al. \cite{anas1998urban}, who proposed models accounting for multiple centers and subcenters. McMillen and Smith \cite{mcmillen2003number} demonstrated that the number of subcenters in an urban area relates systematically to population size and commuting costs.

Importantly for our work, Bertaud \cite{bertaud2003spatial} argued that even polycentric cities typically maintain a dominant center, suggesting that gradient analysis remains valuable in complex urban systems. His concept of ``urban spatial structure'' describes both population distribution and trip patterns within metropolitan areas, providing a bridge between urban form and transport needs.

\subsection{Public Transport Efficiency Metrics and Spatial Determinants}
Public transport system efficiency has been evaluated through various metrics. Vuchic \cite{vuchic2005urban} established fundamental relationships between transport network design and urban form, while Mees \cite{mees2010transport} emphasized the importance of network planning over technological solutions. Newman and Kenworthy \cite{newman1999sustainability} demonstrated strong correlations between urban density and transport energy use across global cities, establishing approximately 35 people per acre as a minimum density threshold for viable public transport service.

Building on this work, Cervero and Guerra \cite{cervero2011urban} quantified the relationship between density and transit ridership, finding that light rail systems become cost-effective at densities of 30 people per acre, while heavy rail requires approximately 45 people per acre.

\subsection{Remote Sensing for Urban Transport Analysis}
Remote sensing applications in urban transport analysis have grown significantly. Thakuriah et al. \cite{thakuriah2017big} reviewed applications of geospatial data in transportation planning, while Taubenböck et al. \cite{taubenboeck2012monitoring} used remote sensing to identify urban growth patterns and their implications for sustainable transport planning. Li et al. \cite{li2020deep} demonstrated methods for extracting road networks from high-resolution satellite imagery, and Barrington-Leigh and Millard-Ball \cite{barrington2017world} tracked global street-network connectivity and its relationship to transportation efficiency.

\subsection{Multi-modal Satellite Data Fusion for Urban Studies}
The integration of multiple satellite data types has proven particularly valuable for comprehensive urban analysis. Zhu et al. \cite{zhu2017deep} reviewed methods for fusing optical and SAR data, highlighting complementary strengths that improve urban feature extraction. Esch et al. \cite{esch2013urban} developed the Global Urban Footprint using SAR data, demonstrating its effectiveness for identifying built-up areas globally, while Pesaresi et al. \cite{pesaresi2013global} created similar products using optical imagery through the Global Human Settlement Layer project.

Gamba and Dell'Acqua \cite{gamba2016multi} showed that multi-sensor fusion techniques significantly improve the accuracy of urban area delineation and structural characterization. Li et al. \cite{li2015record} addressed the challenge of identifying urban centers from satellite imagery, developing automated methods that align with our approach to gradient analysis. Similarly, Taubenböck et al. \cite{taubenboeck2019new} proposed methods for classifying urban spatial patterns from remote sensing data that directly inform public transport planning.

Our work builds upon these foundations by integrating density gradient analysis with multi-modal satellite data processing, among other things, for public transport analyses. By quantifying the relationship between urban morphology and transport network requirements through our proposed metrics, we provide a novel computational tool that bridges urban remote sensing and transport planning disciplines.

\section{Methodology}

\subsection{Data Sources and Preprocessing}
Our approach utilizes optical satellite imagery and Synthetic Aperture Radar (SAR) imagery from the Copernicus Data Space \cite{drusch2012sentinel, torres2012gmes, berger2012esa}. The multi-modal nature of our dataset leverages the complementary strengths of both data types: optical imagery provides visual features and textural information, while SAR offers penetration capability and illumination-independent measurements of urban structures \cite{zhu2017deep, esch2013urban}.

For our analysis pipeline, we use:
\begin{itemize}
\item Optical Sentinel-2 satellite true color images (RGB format)
\item SAR Sentinel-1 backscatter intensity images (Interferometric Wide Swath, VV-polarisation)
\end{itemize}

Preprocessing includes:
\begin{itemize}
\item Image registration to ensure spatial alignment between optical and SAR data
\item Standardization to a uniform pixel resolution
\item Normalization of intensity values to comparable ranges
\item Edge preservation to maintain critical urban boundaries during processing
\end{itemize}

\subsection{Urban Structure Analysis Pipeline}
Our methodology follows a systematic workflow as illustrated in Figure 1. The process consists of several key sequential steps:

\begin{figure}[!ht]
\centering
\includegraphics[width=\textwidth]{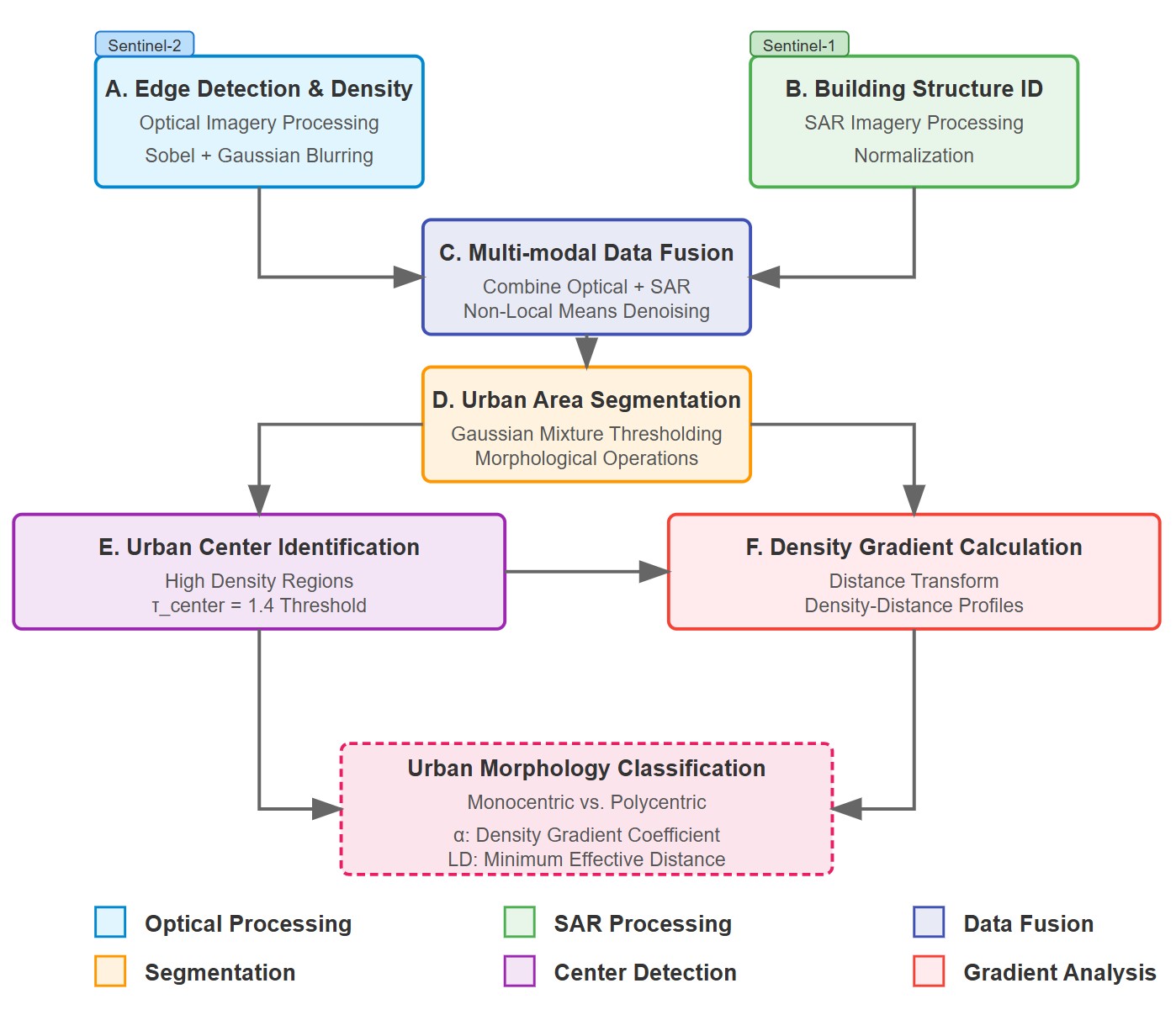}
\caption{Urban Density Gradient Analysis Methodology Pipeline. Flow diagram showing the processing sequence from Sentinel-1/2 imagery through multi-modal fusion, segmentation, and parallel analysis of urban centers and density gradients, yielding classification metrics $\alpha$ and LD for characterizing urban morphology.}
\label{fig:pipeline}
\end{figure}

A. \textbf{Edge detection and density mapping from optical imagery}: We apply Sobel operators to identify structural boundaries in optical images. These edge maps provide critical information about the spatial organization of urban features. Edge density is then calculated using Gaussian blurring to create a continuous density field that highlights areas with concentrated structural elements.

B. \textbf{SAR imagery processing for building structure identification}: SAR backscatter data is particularly sensitive to building structures due to corner reflections and multiple bounces from vertical surfaces \cite{esch2013urban, gamba2016multi}. We normalize the SAR data for integration with optical-derived features to ensure comparable intensity scales.

C. \textbf{Multi-modal data fusion of optical and SAR data}: We combine the edge density information from optical imagery with the structural information from SAR data to create a comprehensive urban density map. To remove noise while preserving important structural edges, we apply Non-Local Means denoising, which is particularly effective at maintaining sharp transitions in urban boundaries.

D. \textbf{Urban area segmentation}: Using density thresholds derived from histogram decomposition of our combined image, we segment the urban landscape into three primary categories. The segmentation is performed using Gaussian mixture model decomposition of the combined image histogram to automatically determine optimal thresholds:

\begin{equation}
S(x,y) =
\begin{cases}
  1 \text{ (Water)}, & \text{if } \rho(x,y) < \tau_{water} \\
  2 \text{ (Terrain)}, & \text{if } \tau_{water} \leq \rho(x,y) < \tau_{urban} \\
  3 \text{ (Urban)}, & \text{if } \rho(x,y) \geq \tau_{urban}
\end{cases}
\end{equation}

Where:
\begin{itemize}
\item $S(x,y)$ is the segmentation class at pixel location $(x,y)$
\item $\rho(x,y)$ is the combined density value at pixel location $(x,y)$
\item $\tau_{water}$ is the water threshold determined as the first intersection point between Gaussian components
\item $\tau_{urban}$ is the urban threshold determined as the second intersection point between Gaussian components
\end{itemize}

These thresholds are automatically calculated for each image through the following process:

\begin{enumerate}[label=\alph*.]
\item Decompose the histogram of the combined image into three Gaussian components
\item Identify intersection points between adjacent Gaussian components
\item Use the first intersection as the water threshold and the second intersection as the urban threshold
\end{enumerate}

This adaptive approach improves robustness across different images and geographic regions by automatically adjusting thresholds to the specific characteristics of each image.

Morphological operations are then applied to refine the urban mask and remove noise:

\begin{equation}
U_{refined} = \text{Close}(\text{Dilate}(U_{initial}, k), k)
\end{equation}

Where:
\begin{itemize}
\item $U_{initial}$ is the initial urban mask where $S(x,y) = 3$
\item $k$ is a $5 \times 5$ structuring element
\item Dilate and Close are standard morphological operations
\end{itemize}

To ensure meaningful urban analysis, we filter out small disconnected patches:

\begin{equation}
U_{final}(x,y) =
\begin{cases}
1 & \text{if } (x,y) \in C_i \text{ and } \text{Area}(C_i) \geq 100 \text{ pixels} \\
0 & \text{otherwise}
\end{cases}
\end{equation}

Where:
\begin{itemize}
\item $C_i$ represents the $i$-th connected component in the urban mask.
\end{itemize}

E. \textbf{Urban center identification}: Urban centers are identified as regions with particularly high density values, defined by:

\begin{equation}
C(x,y) =
\begin{cases}
  1, & \text{if } \rho(x,y) > \tau_{center} \text{ and } U_{final}(x,y) = 1 \\
  0, & \text{otherwise}
\end{cases}
\end{equation}

Where:
\begin{itemize}
\item $C(x,y)$ indicates whether pixel $(x,y)$ is part of an urban center
\item $\rho(x,y)$ is the combined density value
\item $\tau_{center} = 1.4$ is the urban center threshold
\item $U_{final}(x,y)$ is the final urban mask
\end{itemize}

This approach aligns with established methodologies for identifying urban centers from remote sensing data \cite{li2015record, taubenboeck2019new}.

F. \textbf{Distance-based density gradient calculation}: We calculate how urban density changes with distance from identified urban centers using a Euclidean distance transform. For each distance increment, we calculate the mean density of all urban pixels at that distance, creating a density-distance profile that characterizes the urban structure.

To quantify this relationship systematically, we implement the following approach:

\begin{enumerate}[label=\alph*.]
\item \textbf{Distance Transform}: Using the identified urban center points as reference locations, we apply a Euclidean distance transform to create a distance map where each pixel value represents its distance from the nearest urban center:

\begin{equation}
D(x,y) = \min_{(c_x,c_y) \in C} \sqrt{(x-c_x)^2 + (y-c_y)^2}
\end{equation}

Where:
\begin{itemize}
\item $D(x,y)$ is the distance value at pixel location $(x,y)$
\item $C$ is the set of all urban center points (locations where urban density exceeds the center threshold)
\end{itemize}

\item \textbf{Urban Area Filtering}: We extract distances and densities only for pixels within urban areas:

\begin{equation}
D_{urban} = \{D(x,y) \mid (x,y) \in U\}
\end{equation}

\begin{equation}
\rho_{urban} = \{\rho(x,y) \mid (x,y) \in U\}
\end{equation}

Where:
\begin{itemize}
\item $U$ is the set of all urban pixels (as identified in our segmentation)
\item This ensures our gradient analysis focuses only on built-up areas
\end{itemize}

\item \textbf{Distance Binning and Mean Density Calculation}: For each integer distance value, we calculate the mean density of urban pixels at that distance:

\begin{equation}
\rho(d) = \frac{1}{|P_d|} \sum_{p \in P_d} \rho(p)
\end{equation}

Where:
\begin{itemize}
\item $P_d = \{p \in U \mid d \leq D(p) < d+1\}$ is the set of urban pixels with distance in bin $d$
\item Each bin has a width of 1 pixel, which corresponds to the satellite resolution (typically 5-20m)
\end{itemize}
\end{enumerate}

The result is a series of points $(d, \rho(d))$ that represent how urban density changes with distance from urban centers. This density-distance profile forms the foundation for our gradient analysis and urban morphology classification. For subsequent analysis, we convert the pixel distances to real-world units (kilometers) by multiplying by the satellite resolution. The resulting density gradient profile directly informs public transport planners by revealing the spatial distribution of urban density.

\subsection{Density Gradient Metrics}
We calculate two key metrics for each analyzed city:

A. \textbf{Density Gradient Coefficient ($\alpha$)}: The slope of the regression line through local minima in the density-distance curve, measured in units per kilometer. This value indicates how rapidly urban density decreases with distance from centers.

We calculate $\alpha$ through linear regression on selected density minima points using:

\begin{equation}
\alpha = \frac{n\sum_{i=1}^{n}(d_i \cdot \rho_i) - \sum_{i=1}^{n}d_i \sum_{i=1}^{n}\rho_i}{n\sum_{i=1}^{n}d_i^2 - (\sum_{i=1}^{n}d_i)^2}
\end{equation}

Where:
\begin{itemize}
\item $d_i$ is the distance from urban center (in km) at point $i$
\item $\rho_i$ is the urban density value at point $i$
\item $n$ is the number of local minima points identified in the density gradient
\end{itemize}

The $\alpha$ coefficient provides a quantitative measure of urban compactness, analogous to the exponential decay parameter in Clark's urban density model \cite{clark1951urban}, but adapted for multi-modal satellite data. Steeper gradients (more negative $\alpha$ values) typically indicate more compact urban forms with rapid density decreases moving away from centers.

B. \textbf{Minimum Effective Distance (LD)}: The distance at which urban density reaches a target threshold value considered minimal for efficient public transport service. This is calculated as the x-intercept of the regression line at the target density.

We calculate LD using the formula:

\begin{equation}
LD = \frac{\rho_{target} - \beta}{\alpha}
\end{equation}

Where:
\begin{itemize}
\item $\rho_{target}$ is the target density threshold (typically set to the minimum density found in urban areas)
\item $\beta$ is the y-intercept of the regression line
\item $\alpha$ is the density gradient coefficient calculated above
\end{itemize}

This metric directly relates to Newman and Kenworthy's \cite{newman1999sustainability} and Cervero and Guerra's \cite{cervero2011urban} findings on minimum density thresholds for viable public transport. It represents the effective radius within which public transport service is likely to be efficient based on density patterns.

\subsection{Urban Morphology Classification}
We classify cities based on their density gradient plots:

A. \textbf{Monocentric Cities}: Characterized by a single dominant peak in the density-distance curve, with density decreasing relatively uniformly with distance from the center. We aim to extend this definition to include cases with steeper density gradients, where density drops more rapidly as distance from the center increases.

B. \textbf{Polycentric Cities}: Exhibit multiple peaks in the density-distance curve, indicating several urban centers of varying intensity.

The peak detection algorithm identifies local maxima in the density gradient plot with a prominence threshold scaled to the data range. Mathematically, we identify peaks where:

\begin{equation}
\rho_i > \rho_{i-1} \text{ and } \rho_i > \rho_{i+1} \text{ and } \rho_i - \min(\rho_L, \rho_R) > p \cdot (\rho_{max} - \rho_{min})
\end{equation}

Where:
\begin{itemize}
\item $\rho_i$ is the density at point $i$
\item $\rho_L$ and $\rho_R$ are the lowest density values between point $i$ and the nearest higher peaks to the left and right
\item $p$ is the prominence factor (typically 0.02)
\item $\rho_{max}$ and $\rho_{min}$ are the maximum and minimum density values across the entire profile
\end{itemize}

This classification methodology aligns with established urban morphology theories from Bertaud \cite{bertaud2003spatial} and Anas et al. \cite{anas1998urban}, providing a quantitative basis for distinguishing between different urban spatial structures using satellite data. Our approach not only identifies multiple centers in polycentric cities but also quantifies their relative importance and spatial relationships. This information is crucial for designing efficient public transport networks that serve complex urban spatial structures.

\section{Results}

This section presents the results of applying our multi-modal satellite imagery analysis methodology to two Polish urban areas with contrasting morphologies: Malbork and Kłodzko. We present the output from each stage of our processing pipeline, demonstrating the progression from raw satellite data to quantified urban density gradients.

\subsection{Satellite Imagery Results}
In Figure 2, optical and SAR satellite imagery are presented for two analyzed urban areas: Malbork (left column) and Kłodzko (right column). The optical images (Figure 2a-b, top row) reveal distinct urban morphologies. Malbork's optical image shows a compact urban structure divided by the Nogat River, with the distinctive castle complex visible in the central area. The two parts of the city separated by the river differ in size and complexity, with the eastern (right) section displaying a more complex urban pattern. Malbork's road network follows a predominantly radial configuration, with most roads converging toward the city center.

In contrast, Kłodzko's optical image displays an urban area whose development is constrained by valley topography, with settlement distributed along several corridors defined by the surrounding terrain. This distinctive morphology is further shaped by the main road that traverses the urban area from south to north, resulting in a city configuration markedly different from Malbork's radial structure.

The SAR backscatter intensity images (Figure 2c-d, bottom row) highlight vertical structures, with brighter pixels indicating stronger returns from building walls and roofs. These SAR images clearly align with patterns visible in the optical imagery. Malbork's SAR image exhibits concentrated brightness in its central area, corresponding to its dense historic core, while Kłodzko's SAR image reveals multiple bright zones distributed across its urban extent, reflecting its more dispersed development pattern.

\begin{figure}[!ht]
\centering
\includegraphics[width=\textwidth]{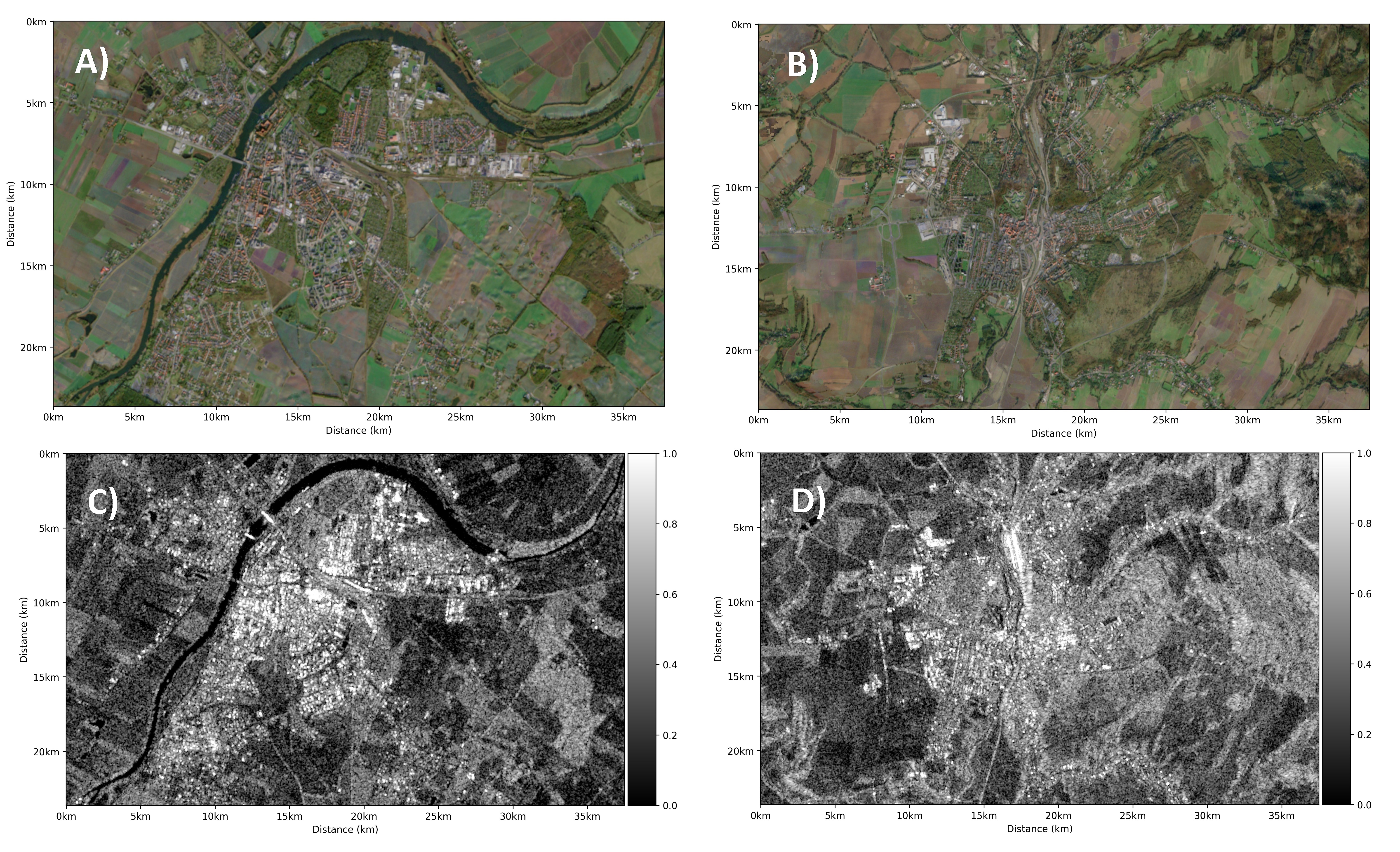}
\caption{Input satellite data: (a) Malbork optical RGB image (left) and (b) Kłodzko optical RGB image (right) in the top row; (c) Malbork SAR backscatter intensity image (left) and (d) Kłodzko SAR backscatter intensity image (right) in the bottom row.}
\label{fig:satellite}
\end{figure}

\subsection{Edge Detection and Fusion Results}
In Figure 3, the edge detection results and combined optical-SAR urban density maps are presented. The top row shows edge detection results for Malbork (Figure 3a, left) and Kłodzko (Figure 3b, right). These edge maps highlight the structural boundaries present in the urban fabric, revealing distinct patterns in both urban areas. Malbork's edge map displays a dense network in the central area with more regular, radially-organized patterns in peripheral zones. Kłodzko's edge map, by contrast, exhibits more irregular structural patterns that reflect the topographical constraints on urban development.

The bottom row of Figure 3 presents the combined optical-SAR urban density maps for Malbork (Figure 3c, left) and Kłodzko (Figure 3d, right). These maps integrate edge information from optical imagery with backscatter intensity from SAR data to produce comprehensive urban density representations. Brighter areas correspond to higher density values, effectively revealing the concentration of urban structures.

Malbork's combined image exhibits a more pronounced contrast in brightness distribution, with clearly defined high-density areas that stand out against the surrounding regions. The central area displays particularly bright values, making it easier to detect high-density urban zones. This distinct brightness pattern facilitates clear identification of both the overall urban area boundary and the city center.

In contrast, Kłodzko's combined image shows a more uniform color distribution across its urban extent, making the detection of high-density areas more challenging. This uniformity reflects Kłodzko's more dispersed urban pattern, where density varies less dramatically across the urban area. Despite this relative uniformity, the combined image still effectively delineates the overall urban area boundary, though with less pronounced internal density variations compared to Malbork.

\begin{figure}[!ht]
\centering
\includegraphics[width=\textwidth]{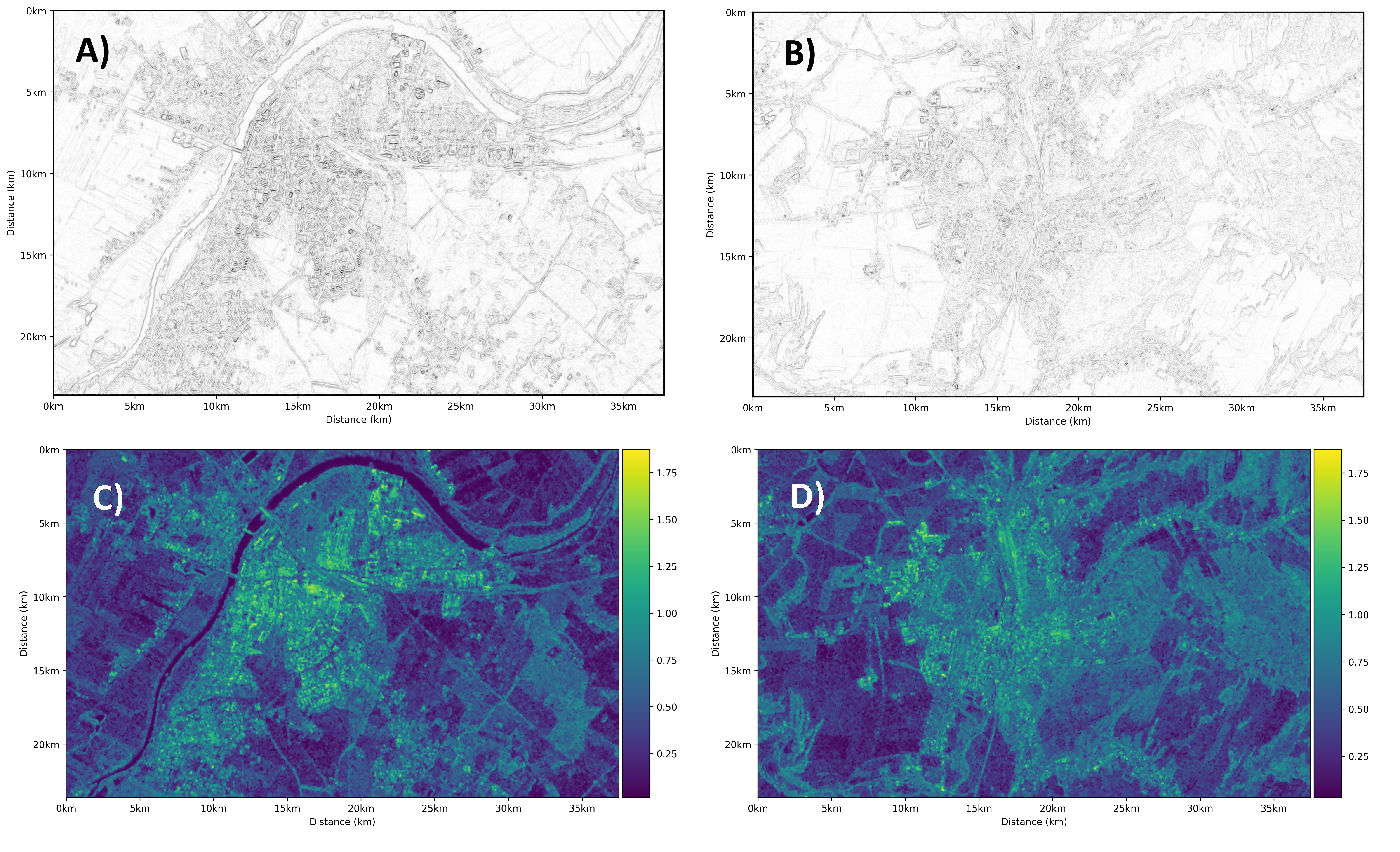}
\caption{Edge detection and fusion results: (a) Malbork edge detection result (left) and (b) Kłodzko edge detection result (right) in the top row; (c) Malbork combined optical-SAR urban density map (left) and (d) Kłodzko combined optical-SAR urban density map (right) in the bottom row.}
\label{fig:edge_fusion}
\end{figure}

\subsection{Histogram Decomposition and Urban Centers}
In Figure 4 there are presented the histogram decomposition results and the segmented urban areas with identified centers. The top row shows histogram decomposition for Malbork (Figure 4a, left) and Kłodzko (Figure 4b, right). In both cases, the algorithm identified three main components in the value histograms.

For Malbork, the histogram decomposition plot reveals three clearly distinct components: the lowest one corresponding to the river values distribution on the combined image, the mid-range component representing land areas, and the third component related to the brightest values indicated as urban. The thresholds were automatically detected at $\tau_{water} = 0.51$ and $\tau_{urban} = 1.02$. It's important to note that the amplitude of these components cannot be treated as a direct measurement of the area of a given landscape type. Rather, it is a value histogram showing the frequency of given intensity values. However, in Malbork's case, a clear intersection of these components arises at the values given above.

In the case of Kłodzko, the situation is quite similar but may be misleading since there are no significant water components in the urban area. Thus, the water threshold ($\tau_{water} = 0.53$) should be treated as distinguishing different land components with varying SAR reflectivity. The urban threshold was detected at $\tau_{urban} = 1.02$, identical to Malbork's value. Despite this complexity, the histogram analysis still effectively enables the delineation of urban areas and the identification of urban centers.

The bottom row of Figure 4 presents the segmented urban areas with identified centers for Malbork (Figure 4c, left) and Kłodzko (Figure 4d, right). These images show urban areas in gray and urban centers in red, based on the density thresholds derived from the histogram analysis. Malbork's segmentation shows a single dominant center corresponding to its historic core, while Kłodzko's segmentation displays multiple proportionate centers distributed across its urban footprint.

\begin{figure}[!ht]
\centering
\includegraphics[width=\textwidth]{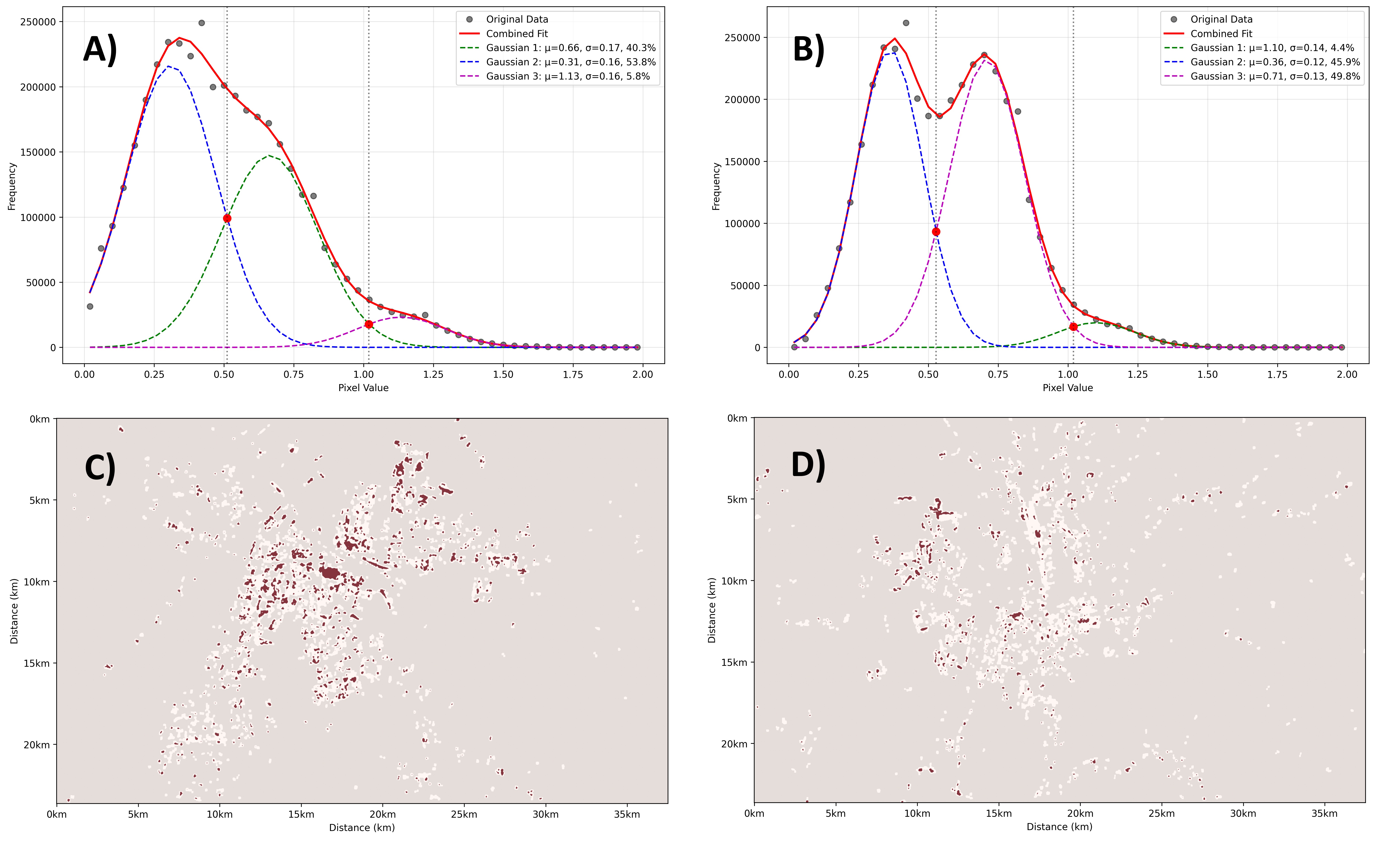}
\caption{Histogram analysis and segmentation: (a) Malbork histogram decomposition (left) and (b) Kłodzko histogram decomposition (right) in the top row; (c) Malbork segmented urban area with identified urban centers (left) and (d) Kłodzko segmented urban area with identified urban centers (right) in the bottom row.}
\label{fig:histogram}
\end{figure}

\subsection{Urban Density Gradient Results}
Figure 5 presents the urban density gradient plots for both study areas, quantifying the spatial distribution of urban density as a function of distance from identified centers. These visualizations enable direct comparative analysis of the fundamental urban structural characteristics of Malbork and Kłodzko.

\begin{figure}[!ht]
\centering
\includegraphics[width=\textwidth]{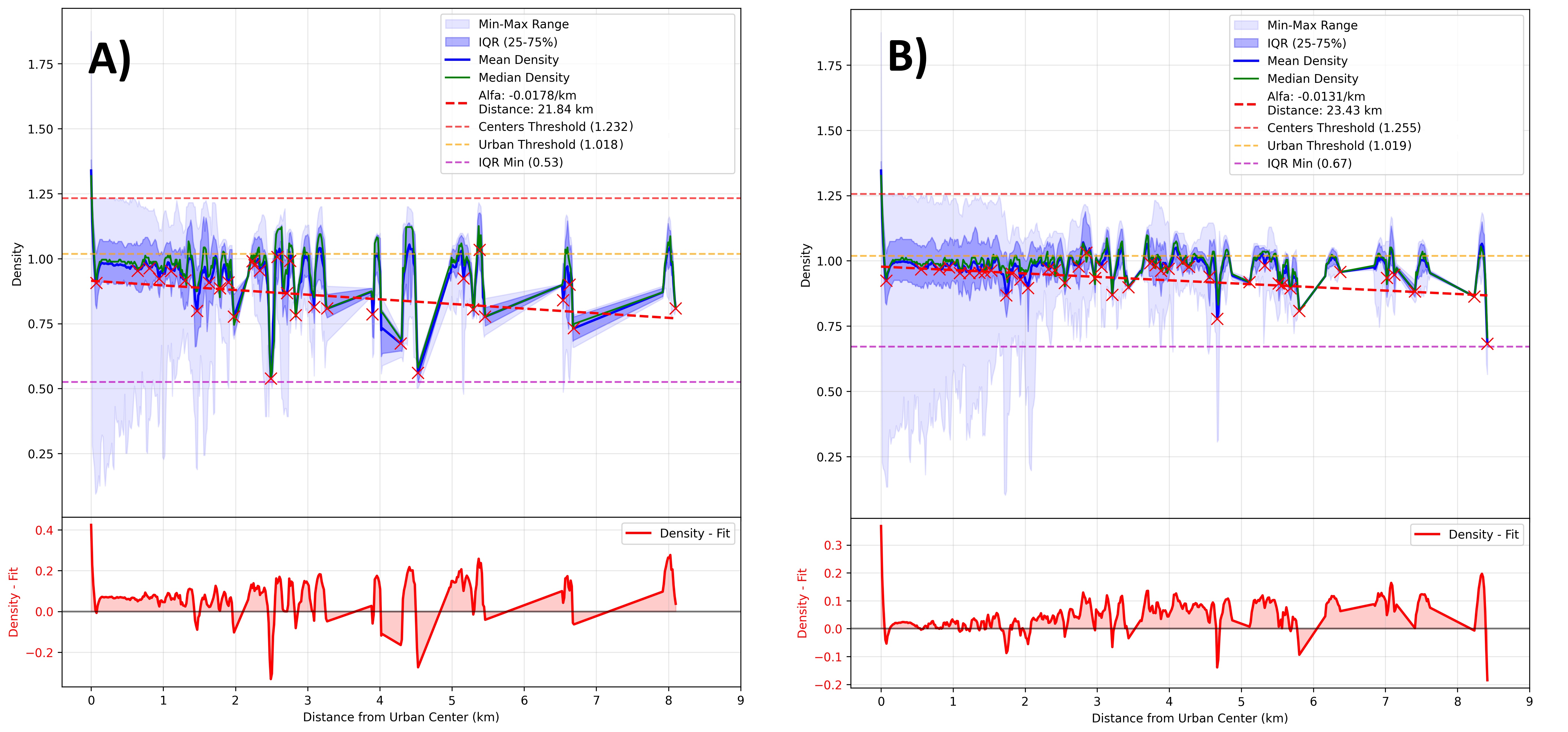}
\caption{Urban density gradient analysis: (a) Malbork density gradient plot (left) and (b) Kłodzko density gradient plot (right). Each figure consists of two connected plots: the upper portion shows gradient density and fitting (yielding $\alpha$ values), while the lower portion shows deviations from the fitted line, highlighting areas of varying uniformity.}
\label{fig:gradient}
\end{figure}

Figure 5a (left) depicts Malbork's density gradient, characterized by a relatively uniform density decrease with distance from the center. The linear regression analysis yields a gradient coefficient $\alpha = -0.018/km$, while the minimum effective distance (LD) is calculated at 21.8 km. Conversely, Figure 5b (right) illustrates Kłodzko's density gradient, with a linear regression yielding $\alpha = -0.013/km$ and LD = 23.4 km. The visualization incorporates additional statistical parameters through color-coding, including the variation of gradient values with respect to all identified centers and the interquartile ranges (IQR, 25-75\%) of all center-to-urban gradient distributions, providing comprehensive quantitative metrics for structural comparison.

The lower portions of Figures 5a and 5b display the difference plots, constructed by calculating the deviation between observed density values and the fitted regression model at each distance increment. This deviation analysis, computed as (observed density - fitted density), effectively identifies regions where urban development diverges from the idealized linear gradient model, revealing areas of both uniformity and non-uniformity in the urban fabric and highlighting zones containing multiple density peaks.

Malbork's difference plot exhibits a notably uniform pattern extending approximately 1 km from the center, with minimal deviation from the regression model, followed by increased variability characterized by multiple peaks of varying amplitudes and inter-peak distances. In contrast, Kłodzko's difference plot demonstrates perturbations throughout the entire urban extent, including areas proximal to the center, with both perturbation frequency and amplitude increasing proportionally with distance. This quantitative analysis confirms the qualitative observations from the satellite imagery examination, indicating that Kłodzko exhibits a less centralized and more heterogeneously distributed urban pattern compared to Malbork's more concentrated structure.

\section{Discussion}

Our study intentionally selected two Polish urban areas—Malbork and Kłodzko—as they represent distinct urban morphologies while sharing comparable regional contexts. This selection enables direct comparative analysis of how different urban forms influence density gradients and their implications for public transport.

Malbork (population: approximately 38,500) is a historically significant city in northern Poland characterized by a classic monocentric urban structure. Its development has been shaped by its medieval origins, with the UNESCO World Heritage Teutonic Knights' castle forming a clear historical center \cite{unesco2011castle}. The city's urban fabric radiates outward from this central point, following a relatively uniform pattern typical of traditional European urban development. With an area of approximately 17.2 km², Malbork maintains a moderate population density of about 2,240 people per km².

In contrast, Kłodzko (population: approximately 27,000) in southwestern Poland exhibits a more complex polycentric structure. Its development has been significantly constrained by the surrounding mountainous topography of the Kłodzko Valley \cite{latocha2017geomorphologically}. The city's urban pattern follows several corridors defined by the valleys, resulting in a more distributed settlement pattern. With an area of approximately 25 km², Kłodzko has a lower population density of about 1,080 people per km², reflecting its more dispersed urban form.

Both study areas were analyzed using Copernicus Data Space satellite imagery at an identical 5-meter spatial resolution. This consistent resolution is sufficient to identify individual buildings and street patterns while enabling efficient processing of the entire urban area. The Copernicus database provided high-quality, cloud-free optical imagery and corresponding SAR data for both locations, making them ideal candidates for our multi-modal analysis approach. The selection of these contrasting urban morphologies—one compact and monocentric, the other dispersed and polycentric—within the same national context provides an excellent test case for evaluating the effectiveness of our density gradient methodology for public transport assessment.

In this study, we used building density as an input parameter, primarily due to its data availability – it might be reliably and efficiently extracted from satellite imaging, which offers wide coverage and regular updates. In contrast, population density data is often outdated, inconsistently collected, or unavailable for many regions, especially in rapidly growing urban areas. Furthermore, building density serves as a reasonable and accessible proxy for population density, as numerous studies have shown a strong statistical correlation between the two. Areas with high building density typically correspond to zones with higher concentrations of inhabitants, making it a practical substitute in spatial planning contexts. By using building density, we ensured a more scalable and reproducible approach to transport planning, especially in regions lacking detailed demographic data. This method also supports dynamic analysis as satellite imaging data can reflect urban development over time.

The density gradient difference plots (Figure 5) reveal distinctive patterns that quantitatively confirm the qualitative observations of urban morphology evident in the satellite imagery. These differences have significant implications for public transport in each city which are revealed or more detaily analized in Figure 6.

\begin{figure}[!ht]
\centering
\includegraphics[width=\textwidth]{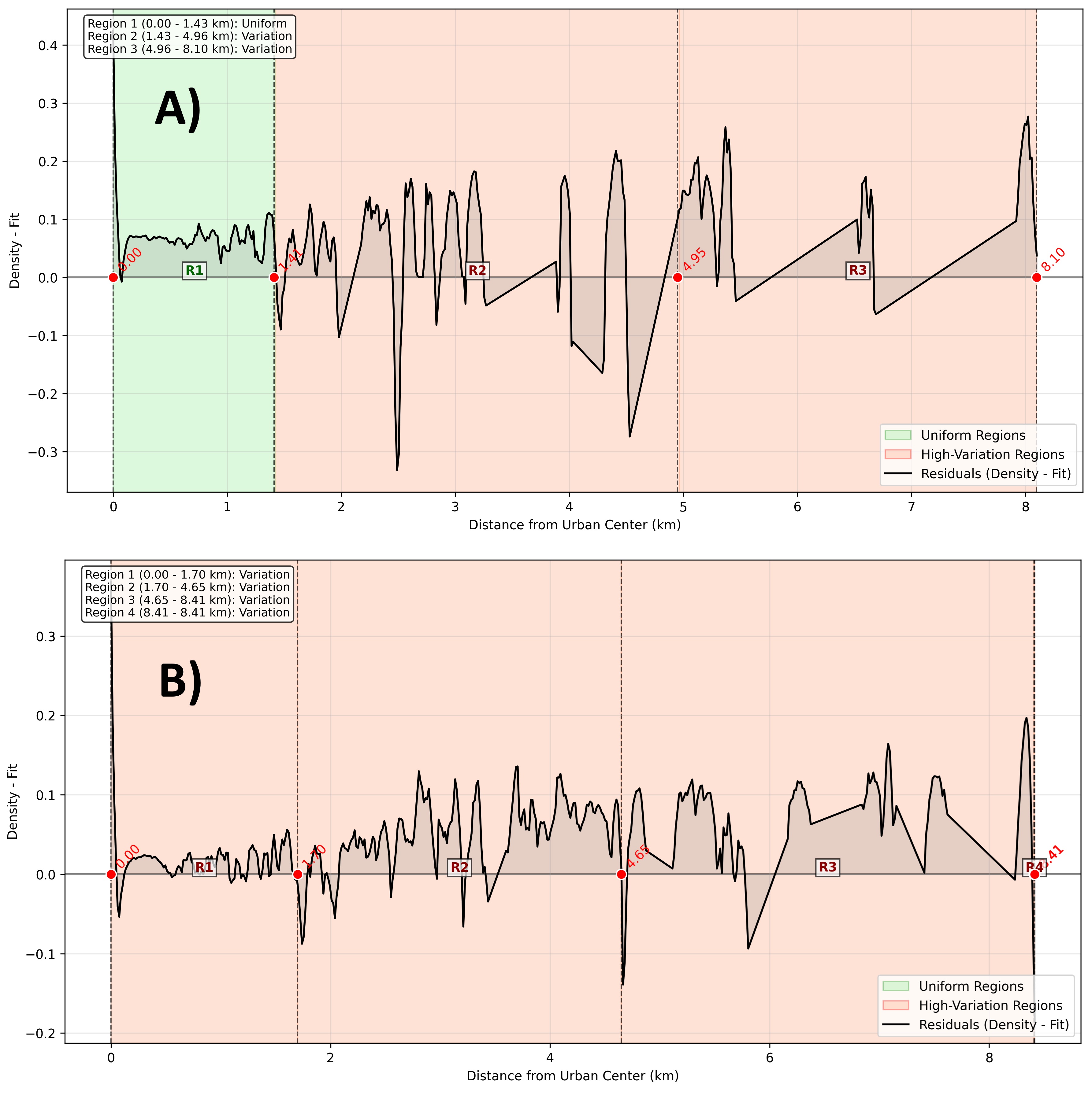}
\caption{Regional analysis of urban density gradient differences: (a) Malbork with three distinct regions (left) and (b) Kłodzko with three regions (right). Green background indicates uniform regions, and red background highlights highly variable regions. For Malbork, Region 1 (0-1.41 km) shows remarkable uniformity, while Regions 2 (1.41-4.96 km) and 3 (4.96-8.10 km) display increasing variability. For Kłodzko, all regions—Region 1 (0-1.70 km), Region 2 (1.70-4.65 km), and Region 3 (4.65-8.41 km)—display significant variability with no uniform central region.}
\label{fig:regional}
\end{figure}

The detailed analysis was performed using a K-means clustering approach applied to the residuals data. This algorithm considers both residual values (density - fit) and their gradients (rate of change), along with shifted versions of both to better capture transition patterns. For each city, we tested different numbers of clusters (between 1 and 3 regions) and selected the optimal segmentation based on minimizing within-region variance while maximizing the number of distinct regions. Small regions (less than 5\% of total data points) were merged with adjacent regions for stability. Each identified region was then classified as either 'uniform' (where standard deviation is less than 70\% of the overall standard deviation) or 'variation' (areas with higher fluctuations in density). This data-driven approach ensures objective identification of urban structural patterns without relying on subjective visual interpretation.

As a result of this analysis, distinct regional characteristics emerged for each city. Malbork's difference plot exhibits a remarkably uniform pattern extending approximately 1.4 km from the center (Region 1 in Figure 6a, highlighted in green), with minimal deviation from the regression model. This uniformity indicates a well-defined, coherent urban core with consistent density—a characteristic ideal for hub-based public transport systems. Beyond this 1.4 km radius, the plot begins to show a variable region (Region 2, 1.4-4.96 km) of multiple medium-height peaks, representing secondary density nodes at intermediate distances. At farther distances (Region 3, 4.96-8.1 km), the peaks demonstrate higher amplitude but wider spacing, indicating sparse but concentrated satellite developments. The entire variable region spans from 1.4 to 8.1 km, as clearly visualized in Figure 6a. This pattern aligns with Malbork's visual appearance in both optical and SAR imagery (Figures 2a, 2c), which shows a compact central area surrounded by distinct, separated development clusters.

In contrast, Kłodzko's difference plot shows perturbations throughout the entire urban extent, including areas proximal to the center. As illustrated in Figure 6b, Kłodzko exhibits a completely variable pattern with no uniform region, with the variable region extending from 0 to 8.4 km. The analysis reveals three distinct regions: Region 1 (0-1.7 km), Region 2 (1.7-4.65 km), and Region 3 (4.65-8.4 km), all characterized by significant perturbations. Both perturbation frequency and amplitude increase proportionally with distance from the identified centers. This pattern indicates a less centralized, more heterogeneously distributed urban structure with multiple density nodes of comparable significance. The consistent presence of peaks even near the nominal "center" suggests that Kłodzko lacks a dominant central core, reflecting its development constraints imposed by valley topography. These observations are consistent with the optical and SAR imagery analysis (Figures 2b, 2d), which reveals a more dispersed urban pattern following topographical features.

The gradient coefficient ($\alpha$) values provide further quantitative evidence of the morphological differences between the two cities. Malbork's steeper gradient ($\alpha = -0.018/km$) compared to Kłodzko's more gradual decline ($\alpha = -0.013/km$) confirms that Malbork has a more centralized urban structure with density decreasing more rapidly with distance from the center. This steeper gradient typically indicates urban forms that can more efficiently support traditional hub-and-spoke public transport networks \cite{newman1999sustainability, cervero2011urban}.

The minimum effective distance (LD) metric further illustrates the transport planning implications of these different urban forms. Malbork's LD value of 21.8 km compared to Kłodzko's 23.4 km indicates that Kłodzko requires a larger service area to effectively cover its population due to its more dispersed development pattern. This finding aligns with established research showing that more compact urban forms generally enable more efficient public transport service coverage \cite{cervero2011urban, kenworthy1999patterns}.

Our analysis revealed potential methodological limitations that should be addressed in future studies. We found that the gradient and density metrics may vary depending on the threshold definitions derived from histogram decomposition. This sensitivity suggests that adding additional channels to the combined image could increase value resolution, resulting in more precise decomposition and ultimately better-defined gradient factors. While our current segmentation algorithms yield consistent results that align with the literature, further refinements could improve accuracy and robustness. Particularly, enhanced edge detection techniques and more sophisticated multi-modal fusion approaches could better delineate urban boundaries in complex topographical settings like those found in Kłodzko.

The validity of our data is further supported by consistency across multiple Polish cities beyond those presented in this study. However, we recognize that threshold sensitivity could introduce variability when applying our methodology to dramatically different urban morphologies or in regions with significantly different built environment characteristics. Future research should explore adaptive thresholding techniques that automatically adjust to regional architectural and urban planning differences.

Our findings suggest promising directions for future research, particularly in developing optimization algorithms for urban transportation systems based on the parameters identified in this study. Since the $\alpha$ coefficient, LD metric, and region characteristics demonstrate consistency across our analysis, they could serve as input variables for models that define optimal urban transportation systems. This approach is planned for our next research phase, along with more comprehensive analysis of a larger sample of cities.

Even with the current data, our urban morphology classification provides clear guidance for public transportation strategies. For Malbork, the well-defined uniform Region 1 indicates an ideal environment for a centralized hub-based transit system, while the surrounding variable regions would benefit from feeder services connecting to this hub. In contrast, Kłodzko's lack of a uniform central region and its three distinct variable regions suggest the need for a distributed transit system with multiple interconnected routes rather than a single dominant hub. This three-tiered communication scheme would better serve Kłodzko's dispersed urban pattern and accommodate its topographical constraints.

\section{Conclusions and Future Work}

Our research demonstrates that multi-modal satellite imagery analysis effectively informs urban planning through density gradient quantification. By combining optical and SAR data, we've developed a methodology that segments urban areas, identifies centers, and calculates key morphological metrics. The density gradient coefficient ($\alpha$), minimum effective distance (LD), and K-means identified regions (R1-R3) provide powerful tools for understanding urban structure and infrastructure requirements.

The comparative analysis of Malbork and Kłodzko reveals how different morphologies demand distinct strategies. Malbork's monocentric structure with a uniform central region ($\alpha = -0.018/km$) suggests an ideal environment for hub-based transit, while Kłodzko's polycentric pattern ($\alpha = -0.013/km$) requires a distributed network with multiple interconnected routes.

Despite promising results, our approach would benefit from enhanced image channel integration. Future work could incorporate temporal analysis, develop predictive models based on our density metrics ($\alpha$, LD) and region classification (R1-R3), and minimize dependence on external socioeconomic data. Optimization algorithms using these parameters could define urban transportation systems tailored to specific morphologies without requiring traditional surveys.

This methodology offers planners a cost-effective, globally applicable approach using freely available satellite data. In an era of rapid urbanization, our work bridges remote sensing and urban planning, supporting more sustainable, efficient, and equitable urban systems worldwide.

\section*{Acknowledgments}

The authors acknowledge the use of 2024/2025 data from the Sentinel-1/Sentinel-2 missions, made available through the European Union's and the European Space Agency's (ESA) Copernicus Programme. The data were accessed via the Copernicus Browser and processed by the authors.

\end{document}